\documentclass[utf8]{article}
\usepackage{url,hyperref,lineno,microtype,subcaption}
\usepackage[onehalfspacing]{setspace}

\usepackage[utf8]{inputenc} 
\usepackage[T1]{fontenc}    
\usepackage{hyperref}       
\usepackage{url}            
\usepackage{booktabs}       
\usepackage{nicefrac}       
\usepackage{microtype}      
\usepackage{multirow}       
\usepackage{xspace}         
\usepackage{natbib}
\usepackage{doi}

\usepackage{tikz}
\usetikzlibrary{bayesnet}
\usepackage{amsmath}
\usepackage{graphicx}
\usepackage{centernot}

\usepackage{xcolor}
\usepackage{soul}

\newcommand{\todoinline}[1]{\hl{TODO: #1}}

\newcommand{\sv}[1]{#1}
\newcommand{\vv}[1]{\boldsymbol{#1}}
\newcommand{\srv}[1]{\mathrm{#1}}
\newcommand{\vrv}[1]{\mathbf{#1}}

\newcommand{\vrx}{\vrv{x}}
\newcommand{\vrz}{\vrv{z}}
\newcommand{\vru}{\vrv{u}}
\newcommand{\vrw}{\vrv{w}}
\newcommand{\vrvv}{\vrv{v}}

\newcommand{\vvx}{\vv{x}}
\newcommand{\vvz}{\vv{z}}
\newcommand{\vvu}{\vv{u}}

\newcommand{\vvw}{\vv{w}}

\newcommand{\srx}{\srv{x}}
\newcommand{\srz}{\srv{z}}
\newcommand{\sru}{\srv{u}}
\newcommand{\srw}{\srv{w}}
\newcommand{\sry}{\srv{y}}
\newcommand{\srvv}{\srv{v}}

\newcommand{\srpred}[1]{\hat{\mathrm{#1}}}
\newcommand{\spred}[1]{\hat{#1}}

\newcommand{\srpy}{{\srpred{y}}}

\newcommand{\spredy}{\spred{y}}

\newcommand{\zmapx}{\alpha}
\newcommand{\xmapz}{\zmapx^{-1}}

\newcommand{\dset}{\mathcal{D}}

\newcommand{\ind}{{\perp\!\!\!\!\perp}}
\newcommand{\delete}[1]{}

\newcommand{\concept}[1]{\textsf{#1}\xspace}
\newcommand{\young}{\concept{Young}}
\newcommand{\gray}{\concept{Gray Hair}}
\newcommand{\gender}{\concept{Gender}}
\newcommand{\glasses}{\concept{Glasses}}
\newcommand{\makeup}{\concept{Makeup}}
\newcommand{\attractive}{\concept{Attractive}}
\newcommand{\smiling}{\concept{Smiling}}
\newcommand{\bignose}{\concept{Big Nose}}

\DeclareMathOperator{\dop}{do}

\usepackage{titlesec}
\titleformat{\paragraph}[runin]
  {\normalfont\bfseries} 
  {}                    
  {0pt}                 
  {}
\titlespacing*{\paragraph}{0pt}{3.25ex plus 1ex minus .2ex}{1em}

 \tikzset{intervened-s/.style= {dot, thick, draw, circle, double, fill=black, minimum size=0.2cm, font={\footnotesize}}}

\usetikzlibrary{arrows,plotmarks,decorations.markings,trees,shapes,pgfplots.groupplots,automata,circuits}
 \tikzset{dot/.style = {circle, fill, minimum size=#1,inner sep=0pt, outer sep=0pt, fill, circle},dot/.default = 6pt}
 \tikzset{dot2/.style = {circle, fill, color=black!40,minimum size=6pt,inner sep=0pt, outer sep=0pt, fill, circle}}
 \tikzstyle{a}=[->,>=stealth,dashed]
 \tikzstyle{a2}=[->,>=stealth]
 \tikzstyle{nodo}=[ellipse,draw=black!100,fill=black!0,line width=.7pt,minimum width=1.2cm,minimum height=0.8cm,text width=1.2cm,text centered]
 \tikzstyle{nodo2}=[ellipse,draw=black!100,fill=black!10,line width=.7pt,minimum width=1.2cm,minimum height=0.8cm,text width=1.2cm,text centered]
 \tikzstyle{nodo3}=[ellipse,draw=black!100,fill=black!30,line width=.7pt,minimum width=1.2cm,minimum height=0.8cm,text width=1.2cm,text centered]
 \tikzstyle{arco}=[draw=black!80,line width=.7pt, postaction={decorate}, decoration={markings,mark=at position 1.0 with {\arrow[ draw=black!80,line width=.7pt]{>}}}]

\title{A Framework for Causal Concept-based\\ Model Explanations}

\begin{document}
\author{
Anna Rodum Bjøru  \and Jacob Lysnæs-Larsen 
\and Oskar Jørgensen  \and Inga Strümke \and 
Helge Langseth
}
\date{Norwegian University of Science and Technology}

\maketitle

\begin{abstract} 

This work presents a conceptual framework for causal concept-based post-hoc Explainable Artificial Intelligence (XAI), based on the requirements that explanations for non-interpretable models should be understandable as well as faithful to the model being explained. Local and global explanations are generated by calculating the probability of sufficiency of concept interventions. Example explanations are presented, generated with a proof-of-concept model made to explain classifiers trained on the CelebA dataset. Understandability is demonstrated through a clear concept-based vocabulary, subject to an implicit causal interpretation. Fidelity is addressed by highlighting important framework assumptions, stressing that the context of explanation interpretation must align with the context of explanation generation.
\end{abstract}


\section{Introduction}
\label{sec:intro}

With state-of-the-art machine learning models used for an increasing variety of high stakes decision making, a well-documented challenge is the lack of transparency introduced by the complexity of the models employed \citep{Goodman_Flaxman_2017,gerlings2021}. The field of Explainable Artificial Intelligence (XAI) aims to ensure models are equipped with the ability to provide explanations for their predictions. One way to approach the task of explaining model behaviour is to design methods that generate explanations for a given pre-trained, black-box model. This is referred to as post-hoc XAI \citep{barredoarrieta2020}, and an obvious advantage of this approach is that a system's ability to explain its behaviour is treated independently of optimising model performance. 

XAI explanations are expected to meet certain criteria. It is critical that 1) explanations are understandable to the explainee \citep{miller19}, and that 2) explanations are faithful to the explanandum, i.e. the model \citep{jacovi2020}. Understandability is receiver-dependent, but generally requires meaningful content and a concise presentation. Fidelity is evaluated with respect to the model to be explained, and any explanation must accurately reflect the model's behaviour. This includes no presence of bias, which for instance could be the result of strategically restricting the scope of explanation, etc. 

While understandability and fidelity are independent qualities of explanations, maximising each in isolation is not trivial. For instance, a challenge to be considered is that explanations that align with what is expected, may be perceived as more understandable. This introduces a preference for explanations that confirm prior beliefs, which results in bias if allowed to influence explanation selection. Moreover, a faithful, unbiased explanation that is misinterpreted will equally mislead the explainee.

This work presents a causal XAI framework for post-hoc explanation of arbitrary models, aiming to facilitate both understandability and fidelity within a defined context. 

First, understandability is addressed through the argument that explanations should be causal in an understandable vocabulary. The requirement for explanations to be causal is based on this being argued the implicit structure of explanation employed in human communication \citep{miller19}. Thus, ensuring causally valid explanations aids in aligning intended interpretation with actual interpretation. This in turn eliminates the need for additional instructions on how to interpret the explanation, allowing a more succinct representation. 

The requirement for understandable vocabulary emphasises the likely discrepancy between the set of data features that the model takes as input, and the set of features ideal for explanation.
This distinction is the key foundation of a group of XAI methods collectively known as Concept-based XAI (C-XAI) \citep{poeta2023}. The central assumption of this approach is that while models are equally capable of processing information in the form of pixels, graphs and tokens, human communication typically assumes a vocabulary consisting of units corresponding to high-level, semantically meaningful concepts. 

Second, fidelity is addressed through identifying and minimising bias introduced by the explanation generation process. Assumptions made prior to generation of explanations are discussed, in order to provide a well-defined context for explanation. This includes details regarding the comprehensiveness of the vocabulary and the choice of causal model, and how the scope of explanation in turn is influenced.

The final framework admits causal reasoning about model behaviour, including generation of counterfactual events as contrastive explanations. Local explanations are generated counterfactually, relating change to probability of sufficiency. Global explanations are generated both with counterfactuals and with interventional queries.

The contributions of the paper are the following:

\begin{itemize}

    \item A theoretical XAI framework for post-hoc, causal explanation generation is presented (Section \ref{sec:ccxai_framework_overview})
    \item 
    Both concept attributions and counterfactually sound contrastive concept-based explanations are defined based on counterfactual calculation of the probability of sufficiency (Section \ref{sec:causal_xai_explanations})
    \item A proof of concept post-hoc explanation model is detailed and applied to generate local and global explanations for classifiers trained on the celebA dataset (Section \ref{sec:examples}). 
    
\end{itemize}

\section{Background}

\subsection{Notation}
Notation $\srv{x}$/$\vrv{x}$ denotes a scalar/set of random variable(s), with element $i$ of $\vrv{x}$ denoted $\srv{x}_i$. 
Notation $\sv{x}$/$\vv{x}$ is used to denote a scalar/set of value(s). The domain of a variable $\srx$ is denoted by $\Omega_{\srx}$. 

\subsection{Causality}
\label{sec:causalbackground}

Relevant background for discussing causal modelling and inference is reviewed here.

\paragraph{Structural Causal Models}
Given a set of variables $\vrvv$ describing a domain, with causal structure represented by graph $\mathcal{G}$.
Counterfactual inference in $\vrvv$ requires a Structural Causal Model (SCM) \citep{galles97,pearl09}. An SCM has components $\vru, \vrvv, \mathcal{F}$, where $\vrvv$ is a set of endogenous variables, $\vru$ is a set of exogenous variables, and $\mathcal{F}$ is a set of deterministic structural functions, one for each endogenous variable.  A structural function $f_{\srvv}(\mathbf{pa}_\srvv) = \srvv $ determines the value of variable $\srvv$ given its parents $\mathbf{pa}_\srvv$ in the causal graph $\mathcal{G}$. The exogenous variables $\vru$ are introduced as root variables in $\mathcal{G}$, such that each $\sru$ is a parent to exactly one endogenous variable $\srvv$. Here, causal sufficiency is assumed ($\sru_i \ind \sru_j$ for all $\sru_i, \sru_j \in \vru, i \neq j$). Such SCMs are called Markovian SCMs \citep{avin2005identifiability}, and do not allow exogenous causal confounding. Note that an endogenous variable $\srvv$ may be modelled without an exogenous parent if $\srvv$ is assumed to be deterministically given by its endogenous parents. 

The distinction between endogenous and exogenous variables is here assumed to be made based on semantic interpretation. All variables in the environment described by $\mathcal{G}$ that represent concepts of known interpretation, are modelled as endogenous variables, regardless of whether it is observed or not. Then, exogenous variables are included where relevant to account for causal influence of origin outside of the model scope. Exogenous variables are thus always unobserved.

The SCM is further rendered probabilistic by introducing a probability distribution $P(\vru)$ over the exogenous variables.
An SCM where all components $\langle\vru, \vrvv, \mathcal{F}, P(\vru)\rangle$ are defined, is referred to as a Fully specified SCM (FSCM) \citep{pearl09, zaffalon2024efficient}. 
If the probability distribution $P(\vru)$ is unknown, the SCM is referred to as a Partially specified SCM (PSCM). The most general PSCM lets each structural function in $\mathcal{F}$ admit every possible function from its endogenous inputs to its output variable, and is referred to as \textit{canonical}.

\paragraph{Counterfactual queries} A counterfactual query combines observed evidence with a hypothetical, retrospective intervention \citep{pearl09}. An intervention, denoted $\dop(\srvv=v)$, assigns a value for a variable while breaking the dependence between the intervened variable and its causal parents, to simulate action, as opposed to change in observation. A counterfactual query is of the form $P(\srw_{\dop(\srvv=v')} = w' | v, w)$ \footnote{A more common notation for a counterfactual variable $\srw$ given intervention $\dop(\srvv=v')$ is $\srw_{\srvv=v'}$. Here, $\dop(\cdot)$ is kept in the subscript to emphasise that the counterfactual value assignment is an intervention.}, read as the probability that variable $\srw$ had taken value $w'$ if variable $\srvv$ had been intervened to take value $v'$, given that the observed event is $\srvv= v,\srw = w$.

\paragraph{Probability of sufficiency}

Given two binary variables $\srvv, \srw$, the Probability of Sufficiency (PS) is given as $P(\srw_{\dop(\srvv=1)} = 1 | \srw=0, \srvv=0)$ \citep{pearl09}, and gives the probability to which variable $\srvv$ intervened to take on value $1$ is sufficient for variable $\srw$ to take on value $1$, given that it is observed that $\srvv = 0$ and $\srw = 0$. 

For binary variables of certain interpretation, the query may be read as the probability of variable $\srvv$ being a sufficient cause of variable $\srw$, implicitly referencing values at 1 as activations. Generalising the probability of sufficiency to all discrete variables as $P(\srw_{\dop(\srvv=v)} = w | \srw=w', \srvv=v')$ with $v \neq v'$ and $w \neq w'$, the query is rather read with explicit reference to the values, i.e. the probability of $\srvv=v$ being a sufficient cause of $\srw = w$ in a context where neither $v$ nor $w$ is observed.






\subsection{Post-hoc XAI}

This section reviews relevant topics concerning post-hoc XAI.

\paragraph{Type of explanation}
Several types of explanations are defined within the field of XAI. These include \textit{feature attributions} \citep{lime, lundberg17}, which are scores assigned for each data feature to quantify the contribution of the feature towards the model's outcome. By a generalised interpretation, this type of explanation extends to concept attributions, where attribution is measured for variables not explicit in the data representation. 

Another type of explanation is the \textit{contrastive explanation} \citep{wachter18}. For a given sample $\vvx$ to be explained, contrasting data samples that are similar to $\vvx$ but that differ in model prediction are presented as the explanation. This type of explanation is also known as a counterfactual explanation in the XAI field, but the term counterfactual in this context has a different meaning than it does in causality. To avoid confusion, contrastive is used whenever it is the XAI type of explanation that is referred to, while counterfactual explanation refers to explanations generated with counterfactual queries as detailed in Section \ref{sec:causalbackground}.

\paragraph{Scope of explanation}
The scope of explanations for XAI range from local to global \citep{angelov21}. Local explanations consider a single data sample $\vvx$, and explain the model's prediction for this sample by considering the model's behaviour in a local neighbourhood around $\vvx$ only. Global explanations instead explain the model by identifying global trends representative of universal model behaviour. Global explanation techniques can be applied to a subgroup of data samples with certain attributes in common, or to infer relationships descriptive of all data samples.


\paragraph{Goal of explanation}  
While the main goal of XAI is to reveal why a model predicts a certain output, the reasons for doing so vary. These include evaluating model performance, increasing user trust in the decision making process, and aiding users in understanding how a decision can be changed in situations where decisions are not permanent. The latter requires a special case of explanation, where the explanation vocabulary is restricted according to \textit{actionability}, i.e. whether a variable may be feasibly intervened upon in a current world state. Explaining under actionability constraints with the goal of changing a model decision is referred to as \textit{algorithmic recourse}. While it requires reliable causal modelling to accurately predict the result of the suggested interventions, it is concerned with the future outcome of actualised interventions only. 

The type of explanation discussed in this paper is intended to be general, with no pre-defined explainee guiding the process of explanation, and with no assumptions made about the performance or deployment stage of the model. This is done to ensure that explanations are faithful to the model, also when the model cannot be assumed to behave as intended or expected. Counterfactual reasoning about retrospective interventions on all variables is permitted, with no requirement for actionability. While the framework discussed may be applied for algorithmic recourse with some adjustments, the details of this is out of scope of this paper. 

\paragraph{Concept vocabulary}
Concept-based explanations use concepts rather than data features as their vocabulary. While there exist multiple definitions of concepts for XAI, it is the type of concept identified as \textit{symbolic concepts} by \cite{poeta2023}, that is referred to here. 
Thus, a concept corresponds to a high-level attribute or characteristic of the data, with semantic meaning as defined by a human. 

Because concepts are not explicitly defined in the data features, it is common to use probing to detect neurons, or combinations of neurons, that best represent target concepts. Such probes are often linear, but may also be non-linear in cases where the concept is not linearly separable. Furthermore, probes can be trained in both supervised and unsupervised manners, where the former requires labelled data of the human-specified concept, and the latter identifies concepts that are important to the model, but not necessarily interpretable to humans. 

To assess the importance of identified concepts, concept attribution methods such as TCAV \citep{kim2018interpretability} quantify the importance of a concept $\srz$, probed in layer $\ell$, for a given output class $\sv{y}$. 
Concepts are learned using linear classifier probes like logistic regression, resulting in the coefficient-vector $\vv{v}_{\srz}$, which is often interpreted as the direction in which the concept is represented. Let $\vv{f}_{\ell}(\vvx)$ be the result of forwarding the input $\vvx$ to the intermediate activation in layer $\ell$, and $g_{\ell,\sv{y}}(\cdot)$ be the remainder of the network, mapping those activations to the logit of class $\sv{y}$. 
\cite{kim2018interpretability} define concept sensitivity as the directional derivative of $g_{\ell,\sv{y}}$ wrt.\ $\vv{f}_\ell(\vvx)$ compared to the the concept direction $\vv{v}_\srz$. 
For an input $\vvx$, the sensitivity score is thus $\boldsymbol{\nabla}_{\!\!\vv{f}_{\!\ell}}\, g_{\ell,\sv{y}} (\vv{f}_{\!\ell}(\vvx))^{\top} \vv{v}_\srz$.
These scores are aggregated over a background dataset, here a subset of the test data $\dset^*$, providing the ratio of positive concept sensitivities as
\begin{equation} \nonumber
    \mathrm{TCAV}_\ell(\sv{y}, \srz, \mathcal{D}^*) = \frac{\sum_{\vvx \in \dset^*} I\left[ \boldsymbol{\nabla}_{\!\!\vv{f}_{\!\ell}}\, g_{\ell,\sv{y}} (\vv{f}_{\!\ell}(\vvx))^{\top} \vv{v}_\srz > 0\right]}{|\dset^*|} \,   ,
\end{equation}
where $I[A]$ is $1$ if $A$ is true and $0$ otherwise.

\section{Related Work}

Several related works discuss post-hoc model explanation with causal models over concepts that are distinct from data features. \cite{parafita19} use distributional causal graphs for casual modelling, with MLPs outputting concept distributions as structural functions. The framework does not allow exact counterfactual inference, and average effect of interventions is rather considered. The paper further includes a conceptual discussion of requirements for a causal generator mapping concepts to data space. 
The methods presented in \citep{dash22, komanduri2024} restrict the process of learning structural functions for an FSCM to observation-based estimation of parameters of a function under model assumptions such as additive noise, Gaussian noise and linearity. The counterfactual context $\vvu \sim p(\vvu|\vvx)$ given a data sample $\vvx$ is sampled via encodings $\vru = q(\vrx)$ learned from observations. The causal generator used by \cite{dash22} is based on a Generative Adversarial Network (GAN) architecture, while \cite{komanduri2024} employ a Diffusion network. 

The present work extends the modelling of structural equations beyond observation-based estimation under general assumptions. Various levels of modelling are considered, ranging from FSCMs designed with clear interpretation under assumptions based on domain knowledge to canonical PSCMs containing all FSCMs consistent with the data. This illustrates both the importance of modelling choices made regarding structural functions, and the potential for partial information in the form counterfactual intervals when complete FSCMs are not available.

\cite{galhotra2021} 
present an XAI method based on calculating probabilities of sufficiency and necessity with FSCMs. Focusing on tabular data, they introduce causal modelling on top of features in data space, assuming these represent high-level concepts, with the option to include further concepts not part of the model input. While only applicable when features are meaningful concepts, this method does not require a causal generator. The method further assumes a preference regarding model output to be explained, such that explanations are framed in terms of which subsets of input values are sufficient to change a model prediction to a more preferable one, or which values are necessary to retain a preferable prediction whenever the outcome of the explanandum is preferred unaltered. 

Moreover, \cite{galhotra2021} include a discussion on bounding probabilities analytically whenever a fully specified SCM is not available. Counterfactual intervals for XAI is further explored in the present work, considering PSCMs for calculating intervals for counterfactual queries. This allows for well defined subgroups of FSCMs to be considered, whenever imperfect domain knowledge is available, potentially reducing intervals compared to analytic calculations.

Probabilities of sufficiency and necessity are also considered for evaluation of XAI methods. \cite{mothilal2021} define the ideal model explanation based on necessity and sufficiency, such that the explanation is a minimal subset of features that is both necessary and sufficient for a model outcome in a given context. With the assumption that the set of variables for explanation is the input features only, and that these are all independent, types of XAI explanations are discussed in comparison. Probability of necessity is related to contrastive explanations by arguing the features changed a necessary cause for original model output. 
Sufficiency is related to feature attribution explanations, in that highly contributing features across contexts will correspond to sufficient features, more likely to cause the output regardless of other feature values. Experiment highlights high variation across methods in practice. 
\cite{pawar24a}, also assuming independent data features, similarly discuss sufficiency and necessity as criteria for existing XAI methods to be evaluated against, and extends this to cover different approaches to neighbourhood generation as basis for explanation.


\cite{beckers22} considers the related setting where the model to be explained predicts the outcome of actions, and thus is assumed to be causally meaningful. Both sufficient explanations and counterfactual explanations 
are defined for this scenario, under the assumption that the model always predict the correct action outcome. The types of explanations are further divided by distinguishing between degrees of causal influence, in order to identify \textit{actual causes} for explanation. 

\section{Method}

\subsection{Causal XAI framework overview}
\label{sec:ccxai_framework_overview}

This section details a framework for generation of causal post-hoc explanations for a model $h$ in a meaningful vocabulary. The framework was first introduced in \citep{bjoru2025} as a means to classify various XAI methods according to their level of causal abstraction. Here, this framework is extended and actualised as a distinct XAI method for explaining black-box models. 

The framework is based on the distinction between the explanation vocabulary, which is a set of meaningful concepts, and the data features that are the input of the model to be explained. 
In total, three model components make up the complete framework: 1) The black-box model $h$ that will be explained, 2) a mapping $\zmapx$ (with inverse $\xmapz$) from the explanation vocabulary concepts to the data representation as input to $h$, and 3) a causal model $M$ over the explanation concepts. The framework includes the following sets of variables: $\vrz$ denotes the explanation concept vocabulary, $\vru$ denotes the exogenous variables associated with $\vrz$ by $M$, $\vrx$ denotes the data features that are the input to the model $h$, $\srpy$ denotes the output of $h$, and finally $\vrw$ denotes any noise or unknown factors of variation in $\vrx$ such that $\vrx$ is deterministically given by $\vrx = \zmapx(\vrz, \vrw)$.

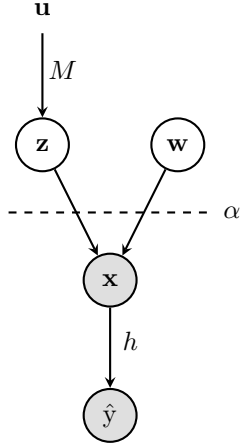
\begin{figure}[ht]
    \centering
    \begin{tikzpicture}[thick, main/.style = {draw, circle}, scale=0.9]
        \node[main, obs,] 
        (yhat) at (0,0) {$\srpy$}; 
        \node[main, obs,]
        (x) at (0,2) {$\vrx$};
        \node[main, latent] (z) at (-1,4) {$\vrz$}; 
        \node[main, latent] (w) at (1,4) {$\vrw$};
        \node[main, draw=none] (u) at (-1,6) {$\vru$};
        \node[main, draw=none] (alpha) at (1.8, 3) {$\alpha$};
        \node[main, draw=none] (h) at (0.3,1.1) {$h$};
        \node[main, draw=none] (M) at (-0.7, 5.1) {$M$};

        \draw[a2] (x) -- (yhat);
        \draw[a2] (z) -- (x);
        \draw[a2] (w) -- (x);
        \draw[a2] (u) -- (z);

        \draw[dashed] (-1.5, 3) -- (1.5, 3);
        
    \end{tikzpicture}
    \caption{The graph details the complete causal framework over variables $\vru\cup\vrz\cup\vrw\cup\vrx\cup\{\srpy\}$ that is presented in this section. Variables $\vrx$ and $\srpy$ are included with gray background, as these are observed variables from the point of view of $h$, the model to be explained. Variables $\vrz, \vrw, \vru$ are generally unobserved.} 
    \label{fig:causalmodel}
\end{figure}

Figure \ref{fig:causalmodel} summarises this framework. The complete model with components $M, \zmapx, h$ can be considered a single causal model over the set of variables $\vru\cup\vrz\cup\vrw\cup\vrx\cup\{\srpy\}$, allowing functions $\zmapx$ and $h$ to be interpreted as representing causal mechanisms. The remainder of this section details each of the model components $h$, $\zmapx$, and $M$ and their respective input and output.

\paragraph{The model $h$ to be explained}

The purpose of the causal framework is to allow explanation of a black-box model $h(\vrx) = \srpy$. This model $h$ takes a set of input features $\vrx$ and predicts a corresponding output $\srpy$. No assumptions are made about the architecture of $h$, and importantly, no assumptions are made about the performance of $h$. 

For $h$ to be interpreted as a causal process, it is important to clearly distinguish prediction $\srpy$ from the attribute $\sry$ being predicted. While the prediction $\srpy$ is always caused by $\vrx$, it is rarely meaningful to model $\vrx$ as a cause of the true attribute $\sry$. Rather, $\sry$ 
can be thought of as a concept that is part of the concept representation $\vrz$ with causal influence on $\vrx$, or otherwise a data feature considered caused by concepts in $\vrz$, and thus correlated with $\vrx$ by confounding.

Explaining $h$ is approached by relating change in $\srpy$ to systematic change in $\vrx$. However, the features $\vrx$ themselves will not in general constitute a useful vocabulary for explanation, as atomic features in $\vrx$ may correspond to pixels in an image, words in a text, nodes in a graph etc. Thus, to quantify meaningful change in $\vrx$, a concept-based vocabulary with known causal structure is considered.

\paragraph{The mapping $\zmapx$ from concepts to data features}

The explanation vocabulary is denoted $\vrz$, with each variable $\srz_i$ representing a meaningful, high-level concept relevant to describe the information represented in $\vrx$. Now $\vrz$ permits explanations in a language understandable for a human explainee. 

In order to describe change in $\vrx$ by change in $\vrz$, an invertible mapping $\zmapx$ between the two sets of variables is required. Ideally, $\vrz$ is sufficient to generate $\vrx$ through a deterministic $\zmapx$, such that all information in $\vrx$ originates from meaningful concepts in $\vrz$ available for explanation. As this is rarely feasible in practice, another set of variables $\vrw$ is introduced, such that $\zmapx(\vrz, \vrw) = \vrx$, with $\vrw$ encoding any unknown factors of variation in $\vrx$ not represented in $\vrz$. Hence, $\zmapx$ is kept a deterministic function. The variables $\vrw$ are not required to be interpretable, and it is assumed that the variable sets $\vrz$ and $\vrw$ are independent. 

The independence $\vrz \ind \vrw$ is assumed in order for the effect of interventions in $\vrz$ on $\vrx$ to be correctly modelled, which is not feasible if $p(\vrw|\dop(\vrz)) \neq p(\vrw)$.
It is noted that $\vrw$ is never to be considered part of the explanation language. For local explanations, $\vrw$ will be considered a context fixed at some value $\vrw = \vvw$, such that the effect of changes to $\vrz$ may be consistently mapped to $\vrx$ through $\zmapx$. 

Another important assumption is thus that $\vrz$, while not a complete model of all information present in $\vrx$, constitutes a sufficiently comprehensive explanation language for $h$'s decision logic to be (partially) exposed. 


Conceptually, $\zmapx$ is considered a decoder $\zmapx(\vvz, \vvw) = \vvx$, and $\xmapz$ is a corresponding encoder $\xmapz(\vvx) = \vvz, \vvw$. The independence $\vrz \ind \vrw$ is assumed preserved by $\xmapz \circ \zmapx$. Dependence among variables in $\vrw$ is irrelevant such that independence here may be enforced if convenient. However, the variables of $\vrz$ can be dependent. 

Note that while $\zmapx$ is modelled as an invertible mapping, it is interpreted as representing a causal process in the data generative direction $\vrz, \vrw \rightarrow \vrx$. The causal influence of a world state on a data instance through the ground truth data generative process is modelled as causal influence of the concepts $\vrz$ as a high level representation of this world state, as well as unknown factors in $\vrw$, on $\vrx$.


\paragraph{The causal model $M$ for causal reasoning}

Expressing explanations in terms of causally meaningful change to $\vrz$ further requires a reliable model of $\vrz$'s causal structure, which represents the underlying physical mechanisms governing the relationships of these concepts in the real world. This structure is represented by a causal graph $\mathcal{G}$. 

\begin{figure}[ht]
    \centering
    \begin{tikzpicture}[thick, main/.style = {draw, circle}]

        \node[label=above:{$\sru_1$}] (u1) at (-1,3) {};   
        \node[label=above:{$\sru_2$}] (u2) at (1,3) {};   
        \node[label=above:{$\sru_3$}] (u3) at (-2,3) {};   
        \node[label=above:{$\sru_4$}] (u4) at (0,3) {};
        \node[label=above:{$\sru_5$}] (u5) at (2,3) {};

        \node[dot,label=right:{$\srz_1$}] (z1) at (-1,1.5) {};   
        \node[dot,label=right:{$\srz_2$}] (z2) at (1,1.5) {};   
        \node[dot,label=right:{$\srz_3$}] (z3) at (-1.7,0) {};   
        \node[dot,label=right:{$\srz_4$}] (z4) at (0,0) {};
        \node[dot,label=right:{$\srz_5$}] (z5) at (1.7,0) {};



        \draw[a2] (z1) -- (z3);
        \draw[a2] (z1) -- (z4);
        \draw[a2] (z1) -- (z5);
        \draw[a2] (z2) -- (z3);
        \draw[a2] (z2) -- (z4);
        \draw[a2] (z2) -- (z5);

        \draw[a] (u1) -- (z1);
        \draw[a] (u2) -- (z2);
        \draw[a] (u3) -- (z3);
        \draw[a] (u4) -- (z4);
        \draw[a] (u5) -- (z5);


        

    \end{tikzpicture}
    \caption{The figure details an example causal graph over concept variables $\srz_1, \srz_2, \srz_3, \srz_4, \srz_5$, each with an exogenous parent $\sru_i$. } 
    \label{fig:exampleG}
\end{figure}
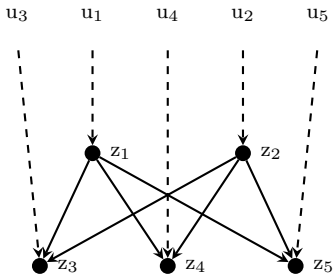

An explicit causal model $M$ is introduced over variables $\vrz$ with structure $\mathcal{G}$. For maximum capacity for reasoning, $M$ is considered an FSCM, and a final set of variables $\vru$ is included, to act as $M$'s exogenous variables. The model $M$ is assumed to be Markovian as described in Section \ref{sec:causalbackground}. Figure \ref{fig:exampleG} presents an example graph $\mathcal{G}$ over variable set $\vrz = \{\srz_i\}_{i=1}^5$, with Markovian exogenous variables $\vru = \{\sru_i\}_{i=1}^5$.

The requirement that $M$ be an FSCM can be challenging to meet in practice, as no amount of data, observational or experimental, will reveal the exact nature of the underlying process to be captured by a structural function. Thus in-depth domain knowledge is typically required for the design of reasonable structural functions, with no automated way of verifying the reliability of a given model. Regardless, an FSCM is required for exact calculation of counterfactual queries, and so is the ideal modelling approach to $M$ for causal explanations. 

If a single, reasonable choice of FSCM cannot be provided, alternatives may be considered, at some cost to explanation quality. Assuming model complexity is sufficiently low, a PSCM may be considered, for which counterfactual queries may be bounded by intervals. The most general PSCM to be considered is referred to as canonical, which provides intervals guaranteed to contain the true counterfactual probability. 

Finally, if no SCM can be provided, $M$ can be modelled as a Causal Bayesian Network (CBN). In this setting, counterfactual reasoning is not feasible, and the framework may be applied for the calculation of interventional queries only.

\vspace{5mm}
\noindent
With $M, \zmapx, h$ as detailed, the causal flow is now summarised. The concept variables $\vrz$ have internal causal structure and are caused by their respective parents among the other concepts, as well as exogenous parents $\vru$, exemplified in Figure \ref{fig:exampleG}. The concepts $\vrz$ together with unknown factors $\vrw$ are causes of the data features $\vrx$ through the data generative process represented by $\zmapx$. Finally, the data features $\vrx$ are causes of the model prediction $\srpy$ through the model $h$. The concepts in $\vrz$ are thus indirect causes of prediction $\srpy$, and this is the relationship explored for explanation. While all variables are part of the extended causal model for explanation, only the variables in $\vrz$ are considered for intervention in order to detect causal relationship. 
Any intervention to variables $\vrx$ or $\srpy$ would remove the link with concept space and the vocabulary chosen for explanation. As $\vru$ and $\vrw$ are unobserved and uninterpretable, interventions here are not informative.

\subsection{Causal XAI explanations}\label{sec:causal_xai_explanations}


This section details how explanations are generated with the presented framework. The main operation of this generation is 
the calculation of probabilities for outcomes of interventions to subsets of concepts in $\vrz$, in a given context.

\subsubsection{Causal calculations}

\newcommand{\intz}[0]{\bar{\vrz}}
\newcommand{\infz}[0]{\bar{\intz}}
\newcommand{\obsz}[0]{\underline{\vrz}}
\newcommand{\intzv}[0]{\bar{\vvz}}
\newcommand{\infzv}[0]{\bar{\intzv}}
\newcommand{\obszv}[0]{\underline{\vvz}}

The framework yields the complete set of variables $\vrx\cup\{\srpy\}\cup\vrz\cup\vru\cup\vrw$. The joint distribution over these variables factorizes as follows: 
\begin{align*}
    p(\vru, \vrw, \vrz, \vrx, \srpy) &= p(\vru)p(\vrw)p(\vrz|\vru)p(\vrx|\vrz,\vrw)p(\srpy|\vrx) \\ &= p(\vru)p(\vrw)p(\vrz|\vru)I[\zmapx(\vrz,\vrw)  = \vrx]I[h(\vrx) = \srpy]
\end{align*}
In order to explain $h$, queries of the generalised form $p(\srpy_{\dop(\intz=\intzv')} | \vv{o})$ are considered, quantifying the effect on the model prediction when intervening on a subset of concepts $\bar{\vrz} \subset \vrz$, conditioned on some observation $\vv{o}$. The observation $\vv{o}$ is the context of the explanation, and different types of explanations are defined based on varying $\vv{o}$.

Notation $\infz$ is used to denote the set of variables in $\vrz$ influenced by an intervention to $\intz$ but not in $\intz$, i.e. the causal descendants of variables $\intz$. Notation $\obsz$ denotes variables in $\vrz$ unaffected by an intervention to $\intz$, such that $\vrz = \intz \cup \infz \cup \obsz$.

\paragraph{Local explanations} 

Local explanations are centred at a single data point $\vvx, \spredy = h(\vvx)$, such that $\vv{o} = \vvx, \spredy$. Having that $h$ is deterministic and that $\vvz, \vvw = \xmapz(\vvx)$, this is equivalent to $\vv{o} = \vvz, \vvw$. The query becomes $p(\srpy_{\dop(\intz=\intzv')} | \vvz, \vvw)$, which is a counterfactual query, calculated as follows:

\begin{align}
    p&(\srpy_{\dop(\intz=\intzv')} | \vvx, \spredy)  \nonumber\\
    &= p(\srpy_{\dop(\intz=\intzv')} | \vvz, \vvw) \nonumber\\ 
    &= \frac{p(\srpy_{\dop(\intz=\intzv')}, \infzv | \obszv, \intzv, \vvw)}{p(\infzv|\obszv, \intzv)} \nonumber\\ 
    &= \frac{1}{p(\infzv|\obszv, \intzv)}\sum_{\vv{u}_{\infz}\in \Omega_{\vru_{\infz}}}  p(\vv{u}_{\infz})p(\srpy_{\dop(\intz=\intzv')}, \infzv | \obszv, \intzv, \vvw, \vv{u}_{\infz})  \nonumber\\ 
    & =  \frac{1}{p(\infzv|\obszv, \intzv)} \sum_{\vv{u}_{\infz}\in \Omega_{\vru_{\infz}}}
    p(\vv{u}_{\infz})p(\infzv| \intzv, \obszv, \vv{u}_{\infz})p(\srpy| \obszv, \vvw, \dop(\intz=\intzv'), \vv{u}_{\infz}) \nonumber\\ 
    & =  \frac{1}{p(\infzv|\obszv, \intzv)} \sum_{\vv{u}_{\infz}\in \Omega_{\vru_{\infz}}}
    p(\vv{u}_{\infz})p(\infzv|\intzv, \obszv, \vv{u}_{\infz} ) \sum_{\infzv^* \in \Omega_{\infz}}I[h\circ \zmapx(\obszv, \intzv', \infzv^*, \vvw)=\srpy]p(\infzv^*| \intzv', \obszv, \vv{u}_{\infz})\nonumber \\
    & = \frac{1}{p(\infzv|\obszv, \intzv)} \sum_{\infzv^* \in \Omega_{\infz}}\sum_{\vv{u}_{\infz}\in \Omega_{\vru_{\infz}}}
     p(\vv{u}_{\infz})p(\infzv|\intzv, \obszv, \vv{u}_{\infz} ) p(\infzv^*| \intzv', \obszv, \vv{u}_{\infz})I[h\circ \zmapx(\obszv, \intzv', \infzv^*, \vvw)=\srpy]
    \nonumber \\
    & = \sum_{\infzv^*\in \Omega_{\infz}}p(\infz^*_{\dop(\intz=\intzv')}|\intzv, \infzv, \obszv)I[h\circ \zmapx(\obszv, \intzv', \infzv^*, \vvw)=\srpy]
    \label{equ:local_counterfactual} 
\end{align} 

\paragraph{Global explanations}

Global explanations are either intervention-based or counterfactual, depending on the explanation context. In the most general case, $\vv{o} = \emptyset$, and the explanation query simplifies as $p(\srpy_{\dop(\intz=\intzv)} |\emptyset) = p(\srpy | {\dop(\intz=\intzv)})$. This query is interventional, and does not require an SCM. Assuming the joint distribution $p(\vrz, \srpy)$ is available, the query is calculated accordingly.

If the context of explanation is a subgroup, the query may become counterfactual. Generally, the context of explanation is $\vv{o} = \vvz_s, \spredy$, where $\spredy$ is a common prediction by $h$ and $\vvz_s$ is a subset of concept values, and the query becomes $p(\srpy_{\dop(\intz=\intzv)} | \vvz_s, \spredy)$.

\begin{align}
\begin{split}
    p(\srpy_{\dop(\intz=\intzv)} | \vvz_s, \spredy) &=     \sum_{\vvx} p(\srpy_{\dop(\intz=\intzv)}| \vvx, \spredy)p(\vvx|\vvz_s, \spredy) \\
    &\approx \sum_{\vvx \in \dset^*} \frac{1}{|\dset^*|}p(\srpy_{\dop(\intz=\intzv)}| \vvx, \spredy)
\end{split}
\label{equ:subgroup_counterfactual} 
\end{align}
The query is estimated using an empirical distribution for $p(\vvx|\vvz_s, \spredy)$, with $\dset^* = \{\vvx: \spredy, \vvz_s\}$ a dataset of samples consistent with the condition $\spredy, \vvz_s$. $p(\srpy_{\dop(\intz=\intzv)}| \vvx, \spredy)$ is calculated as detailed for local explanations.

If the subgroup is not conditioned on a common prediction, i.e. $\vv{o} = \vvz_s$, the explanation query will be counterfactual or interventional depending on whether the intervention $\dop(\intz=\intzv)$ influences any of the context variables or not, respectively. 



\subsubsection{Probability of sufficiency for XAI}

The general probability of sufficiency presented in Section \ref{sec:causalbackground} is adapted and applied for XAI explanations. The target variable of interest is the model prediction $\srpy$. Hypothetical change in the form of a counterfactual intervention is considered for a subset of concept variables $\intz$. The context is either a local instance $\vvx, \spredy, \vvz, \vvw$, or a subgroup defined by common concepts and prediction $\vvz_s, \spredy$ such that $\intz \in \vrz_s$. Then the probability of sufficiency for XAI is given by $p(\srpy_{\dop(\intz = \intzv')} = \spredy' | \vvx, \spredy)$ (local) or $p(\srpy_{\dop(\intz = \intzv')} = \spredy' | \vvz_s, \spredy)$ (subgroup), with $\spredy' \neq \spredy$ and $\intzv' \neq \intzv$. The query quantifies the effect of the intervention $\dop(\intz = \intzv')$ on $\srpy$, relative to observed explanandum $\spredy$ and $\intzv$. 

Note that it is the counterfactual outcome that is the target of the sufficiency query, relative to the observation. The query is thus primarily explanatory with regards to the influence of the changed concepts towards the changed outcome. However, information about the causal influence of the observed concepts on the observed outcome is (partially) revealed, by letting sufficiency of change indicate necessity of the original value. Note that if $|\Omega_{\intz}| > 2$,  a new value $\intzv'$ being sufficient for change to occur in $\srpy$ is not always equivalent to the observed value $\intzv$ being necessary for the observed $\spredy$. Still, the respective scenarios are related.

\subsubsection{Concept attributions}

Given a single data point or a subgroup, the probability of sufficiency can be seen as a measure of attribution for the set of intervened features. The probability of sufficiency is to be read as a measure of attribution of intervened value $\intzv'$ towards alternative outcome $\spredy'$ in a context (e.g. $\vvx, \spredy$), and is only indirectly descriptive of observed $\intzv$'s attribution towards observed $\spredy$. 
In the special case where $|\Omega_{\intz}| = 2$, i.e. it is a single binary concept that is intervened, the probability of sufficiency can be interpreted as a concept importance score as defined for XAI, reflecting the probabilities for the \textit{sufficiency} of the counterfactual concept value towards the counterfactual outcome and the \textit{necessity} of the original concept value towards the original outcome. 



\subsubsection{Contrastive explanations}
\label{sec:contrastive_ccxai}

The presented framework enables the generation of contrastive explanations. The contrastive explanation form considered, based on the explanation form first introduced by \cite{galhotra2021}, is summarised as:

\begin{quote}
    For a data instance $\vvx$ that has concept values $\vvz$, and for which a model $h$ made the decision $\spredy$, the decision would have been $\spredy'$ with probability $p$ had the concept(s) $\intz$ taken value(s) $\intzv'$
\end{quote}
with $\intz$ the subset of the concepts that are changed by (hypothetical) intervention. The probability $p$ is the local probability of sufficiency $p(\srpy_{\intz = \intzv'} = \spredy' | \vvx, \spredy)$. 

When selecting the best contrastive explanations from a set of contrastive samples, similarity with the original data instance is typically measured \citep{wachter18}. When data features are changed independently, similarity can be calculated as number of changed features. With the contrastive form presented here, similarity is interpreted as the number of variables intervened upon, i.e. $|\intz|$. Additionally, the probability $p$ must be considered when evaluating a potential explanation. Generally, it is reasonable to expect $p$ to be high in order for the explanation to be useful. In some cases, a limit $p^*$ for this probability may be given, such that it is required that $p\geq p^*$ for a sample to be considered an explanation. This in turn may allow a simplified explanation form where $p$ is omitted, which is argued helpful in increasing understandability \citep{miller19}.

Searching for optimal contrastive explanations of the form presented, maximising similarity and probability, is left for future work.

\section{Examples}
\label{sec:examples}

\delete{

\textbf{Plan for minimal solution:}

\begin{itemize}
    \item Start with age and best model. Show some anecdotal "proof" of the thing working. Supply image, and proposed counterfactual "If you had gray hair / were using glasses $h$ would think you were young"
    \item Same image: How bad is it with GradCAM?
    \item Calculate the estimated \textit{global} PS for each concept in $M$. Both good and bad model.
Example of interpretation: Out of everyone with young=True and  GrayHair=False, how often do we get young=False when we intervene to have GrayHair=True. Easy for leaf nodes. For Young concept we need to CHANGE young, then do all 4 combos of child-nodes. Anna will aggregate results using the probabilities. Supply results per data-point: z, changed-z, result from h. Discuss what we see here, e.g., Bad model likes gray hair. 
\end{itemize}

Baselines for comparing information-level of explanations:
\begin{itemize}
    \item attribute-level: GradCAM or similar
    \item non-causal aggregation whenever one (leaf-)feature is not enough (intervention in non-leaf has cascading effects). 
\end{itemize}

}

In this section we will exemplify the explanatory process outlined above by way of an example. 
We work with the CelebA dataset \citep{liu2015faceattributes}. This dataset contains more than 200.000 images of celebrities' faces, each annotated with binary attributes describing facial features and accessories (including, e.g., if the subject is \textit{smiling}, \textit{wearing glasses}, of \textit{young age} or is \textit{looking attractive}). 
We will consider two tasks: classifying whether an image is of a \young person and whether the person is \attractive. 

\subsection{Models} 
\label{sec:ex_models}

\begin{figure}[ht]
    \centering

    \begin{tikzpicture}[thick, main/.style = {draw, circle}]

        \node[dot,label=right:\young] (age) at (4,2) {};
        \node[dot,label=left:\gender] (gender) at (0,2) {};
        
        \node[dot,label=below:\concept{Smiling}] (smiling) at (-0.2,0) {};
        \node[dot,label=left:\concept{Makeup}] (m) at (-2,0) {};
        \node[dot,label=below:\concept{5 o'Clock Shadow}] (5) at (2,0) {};
        \node[dot,label=below:\concept{Bushy Eyebrows}] (be) at (5,0) {};
        \node[dot,label=below:\concept{Big Nose}] (bn) at (7,0) {}; 

        \node[dot,label=below:\concept{Arched Eyebrows}] (ae) at (-2,-2) {};
        \node[dot,label=below:\concept{Pointy Nose}] (pn) at (1.2,-2) {};

        \draw[a2] (age) -- (bn);
        \draw[a2] (age) -- (be);
        \draw[a2] (age) -- (5);
        
        \draw[a2] (gender) -- (bn);
        \draw[a2] (gender) -- (be);
        \draw[a2] (gender) -- (m);
        \draw[a2] (gender) -- (smiling);
        \draw[a2] (gender) -- (5);
        \draw[a2] (gender) -- (ae);
        \draw[a2] (gender) -- (pn);
        
        \draw[a2] (m) -- (ae);
        \draw[a2] (m) -- (pn);

    \end{tikzpicture}

    \caption{The causal model over concepts relevant for the attractiveness-classifier} \label{fig:example_attractive}
\end{figure}
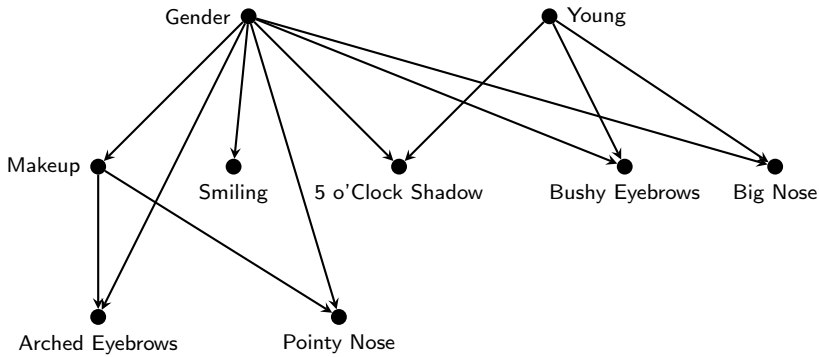

We implemented the conceptual model in Figure \ref{fig:causalmodel} as a computational pipeline, and will now describe each part separately. 
Each image $\vvx_i\in\dset$ in the dataset is annotated with a corresponding set of binary descriptors $\vvz_i$, and we will use a subset of the concepts in $\vrz$ as vocabulary when explaining the classifier $h(\cdot)$.  
For each explanation task, we chose the subset by investigating which concepts correlate the most with the target concept (\young and \attractive). 
The resulting subsets can be identified as the node-sets of the models in Figures \ref{fig:example_young} and \ref{fig:example_attractive}, respectively.

\paragraph{The causal model $M$:}
We build an FSCM over a selected subset of the concepts $\vrz$. 
The causal structure was learned using an expert in the loop-method from pgmpy \citep{Ankan2024}, with a Large Language Model (LLM) acting as the expert deciding on causal direction, and with the forced constraint that concepts \young and \gender should be independent. Conditional independence is also enforced for variables \smiling and \bignose.
The resulting structures are given in Figures \ref{fig:example_young} for the age classification and  Figure \ref{fig:example_attractive} for the classification of attractiveness. 
The parameters of the FSCM defining the probability distribution $p(\vru)$ were chosen to be sparse and meaningful, while compatible with the observed data $\dset$; see \ref{sec:FSCMdetails} for details. PSCM intervals are retrieved using the DCCC method for exact interval calculation \citep{bjoru2025ijar}. Counterfactual queries are calculated using BCAUSE \citep{cabanas2025bcause}.

\paragraph{The classifier $h$:}
We use a 50-layer ResNet following the procedure of \cite{he2015deepresiduallearningimage}. 
While our framework can explain the behaviour of \textit{any} classifier, irrespectively of merit, we note that both classifiers performed at an acceptable level, with accuracies above 80\%. 

\paragraph{The observation model $\zmapx$ and its inverse $\xmapz$:}
In order to have a complete computational pipeline for counterfactual explanations, we train a generative model as a means to represent the mapping between concept-space and data-space. 
In particular, we provide counterfactual data examples as follows:
First, the mapping $\xmapz(\vvx) = \vvz, \vvw$ is partially provided in the dataset itself, as each image $\vvx_i$ is annotated with concepts $\vvz_i$. 
$\vvw$ is intended to capture the remaining information not described by $\vvz$, e.g., image background. 
Here we simply set $\vvw = \vvx$, and will discuss the effects of this simplification in Section \ref{sec:experiment_discuss}.
We aim to evaluate a counterfactual explanation $\dop(\intz=\intzv')$ by calculating $p(\srpy_{\dop(\intz=\intzv')} | \vvx, \spredy)$, cf.\ Section \ref{sec:causal_xai_explanations}. 
This requires generation of counterfactual data-objects $\zmapx(\vvz', \vvw)$ where 
$\vvz'=\{\intzv' \cup \infzv \cup \obszv\}$ is consistent with the causal model $M$, see  Equation \ref{equ:local_counterfactual}.
Our implementation adopts the StarGAN architecture \citep{choi2018starganunifiedgenerativeadversarial} for image-to-image translation in order to provide the functionality of $\zmapx(\vvz',\vvw)$. 
StarGAN takes an image $\vvx$ and a binary vector of high-level concepts $\vvz'$ as inputs and produces an alternation of $\vvx$ consistent with concept description $\vvz'=\{\intzv' \cup \infzv \cup \obszv\}$ but otherwise incentivised to be close to $\vvx$.



\delete{
\begin{itemize}
    \item $\alpha$ - StarGAN - domain translation - should be ok when all relevant domains (young, gender, makeup, gray hair, glasses) are included 
    \item $h$ - classifier for attribute young, binary
    \item Causal graph - LLM for causal direction
    \item CBN - P(child|parents) from data 
    \item FSCM - set of functions to be defined s.t. P(U) given by data
\end{itemize}
}



\begin{figure}[ht]
    \centering

    \begin{tikzpicture}[thick, main/.style = {draw, circle}]




        \node[dot,label=right:{\young}] (age) at (2,2) {};
        \node[dot,label=left:{\gender}] (gender) at (0,2) {};
        \node[dot,label=below:{\gray}] (gh) at (1,0) {}; 
        \node[dot,label=below:{\glasses}] (gl) at (3,0) {};
        

        \node[dot,label=below:{\makeup}] (m) at (-1,0) {};

        \draw[a2] (age) -- (gh);
        \draw[a2] (age) -- (gl);
        \draw[a2] (gender) -- (gh);
        \draw[a2] (gender) -- (m);
        \draw[a2] (gender) -- (gl);


        
    \end{tikzpicture}

    \caption{The causal model over concepts relevant for the age-classifier} \label{fig:example_young}
\end{figure}
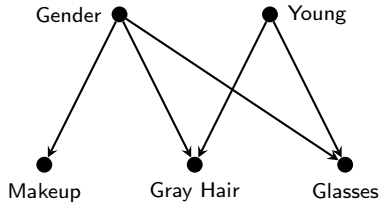

\subsection{Results} \label{sec:results}

\begin{figure}[ht]
    \centering
    \begin{tabular}{c@{}c@{}c@{}c@{}c}
    \includegraphics[width=.175\textwidth]{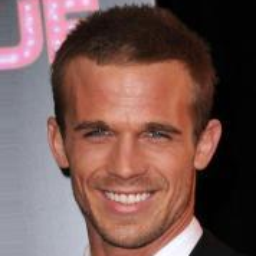}     &
    \includegraphics[width=.175\textwidth]{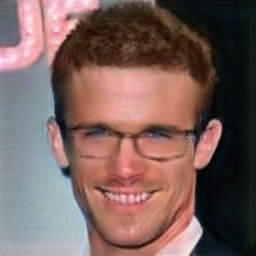}  &
    \includegraphics[width=.175\textwidth]{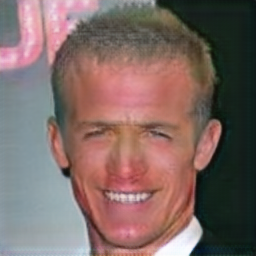} &
    \includegraphics[width=.175\textwidth]{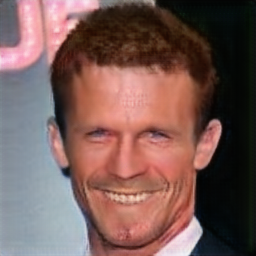}      &
    \includegraphics[width=.175\textwidth]{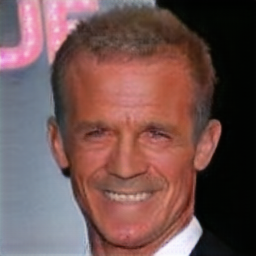} \\ [-4pt]
    \includegraphics[width=.175\textwidth]{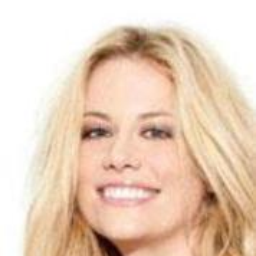}       &
    \includegraphics[width=.175\textwidth]{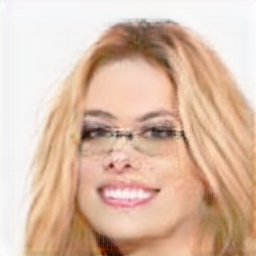}    &
    \includegraphics[width=.175\textwidth]{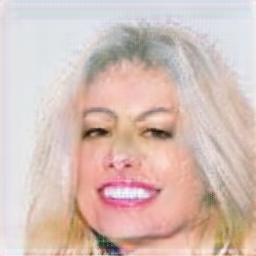} &
    \includegraphics[width=.175\textwidth]{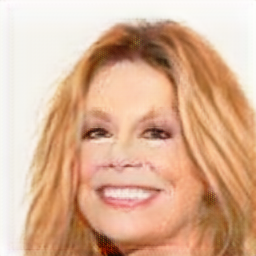}        &
    \includegraphics[width=.175\textwidth]{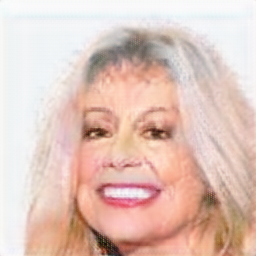}    
    \end{tabular}
    \caption{Example-images from the dataset together with several counterfactual variations. 
    From left to right, the columns contain the original, followed by counterfactuals with $\glasses=1$, $\gray=1$, $\young=0$, and $\gray=1$ {\&} $\young=0$.
    }
    \label{fig:example_face_young}
\end{figure}

\paragraph{Local explanation} 
We first consider the images in Figure \ref{fig:example_face_young}. 
For each row, the original image is given to the left, and we look for explanations of why these people are classified as being young.  
The explanation engine outlined above provides justification by answering counterfactual questions of the type 
\textit{``Would the model $h(\cdot)$ classify the person as young if $\langle$counterfactual$\rangle$?''}, where the counterfactual is defined using the vocabulary shown in Figure \ref{fig:example_young}. 
An intermediate step of producing explanations is to generate counterfactual images as described above. 
The counterfactual images shown in columns 2 and 3 are related to the concepts \glasses and \gray.
Adding glasses is not sufficient for the classifier to change its evaluation of either person in this example, but changing hair colour is sufficient for the male to be classified as old\footnote{We acknowledge that the comparatively poorer quality of the generated images with gray hair, presumably caused by an imbalanced dataset, can be a contributing factor to this result. We discuss this further in Section \ref{sec:experiment_discuss}.}. 

Note that even though \young is a concept in the causal model, its value is in general distinct from the predicted value $\spredy=h(\vvx)$ as the classifier is not infallible. 
Intervening in the causal model $M$ with $\dop(\young=0)$ will imply higher probabilities of both \gray and \glasses  (consider the causal model in Figure \ref{fig:example_young}), in addition to other visual cues related to this concept, like higher affinity for wrinkles and under-eye bags.  
To examine counterfactuals with $\intz=\{\young\}$, we need to marginalize out  the uncertainty in the concepts 
$\infz=\{\glasses, \gray\}$,  cf.\ Equation \ref{equ:local_counterfactual}. 
Two of the counterfactual data-points produced by $\zmapx$ to make this calculation are shown in columns 4 and 5 of Figure \ref{fig:example_face_young}. 
We eventually find that  
$p(\srpy_{\dop(\young=0)}=0 | \vvx, \spredy=1)=1 $ for both subjects, validating the explanation that 
the model $h(\cdot)$ would indeed consider both persons to look old if they were old. 

\begin{figure}[t]
    \centering
    \begin{tabular}{c@{}c@{}c@{}c@{}c}
    \includegraphics[width=.175\textwidth]{figs/87_real.png}        &     
    \includegraphics[width=.175\textwidth]{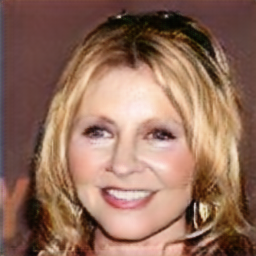}  &     
    \includegraphics[width=.175\textwidth]{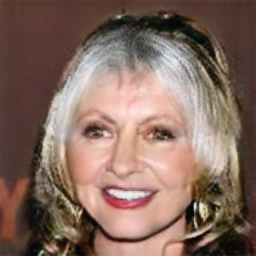}    &     
    \includegraphics[width=.175\textwidth]{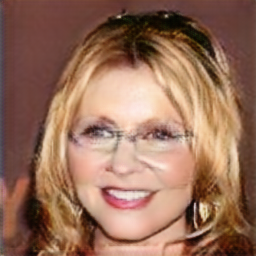} &     
    \includegraphics[width=.175\textwidth]{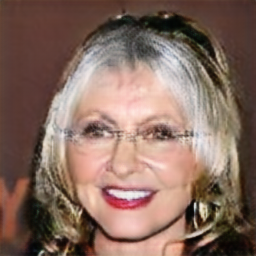}  
    \end{tabular}
 
    \caption{Counterfactual versions of the image to the left (classification results are given in parentheses):
    $\young=0$ ($h(\vvx')$ is young), $\young=0$ {\&} $\gray=1$ (Old), $\young=0$ {\&} $\glasses=1$ (Old), $\young=0$ {\&} $\gray=1$ {\&} $\glasses=1$ (Old)
    }
    \label{fig:example_face_young_fooled}
\end{figure}

We next turn to the images in Figure \ref{fig:example_face_young_fooled}. 
Following from the original image on the left, the figure shows different counterfactual images all consistent with the intervention $\dop(\young=0)$.
The first counterfactual image, obtained by the intervention $\dop(\young=0, \glasses=0, \gray=0)$, is erroneously classified as being of a young person, while the other counterfactual images are correctly identified as being of old subjects. 
Using these counterfactual images to marginalize out the uncertainty over $\infz$, we can calculate $p(\srpy_{\dop(\young=0)} =0| \vvx=\raisebox{-3pt}{\hbox{\includegraphics[height=11pt]{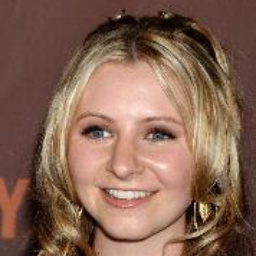}}}, \spredy=1)$, the probability 
that the classifier would classify the subject as old in the  counterfactual world where $\young=0$. 
 The probabilistic model-class employed to do the calculations will provide different results, as showcased in Table \ref{tab:local_ps_young_per_model}.

\begin{table}[t]
    \centering
    \caption{
    Probability of sufficiency for counterfactual intervention on the variable $\intz = \young$ for the leftmost image in Figure \ref{fig:example_face_young_fooled}.
    \textit{FSCM}: The fully specified SCM is the only model representation able to calculate the counterfactual probabilities directly; 
    \textit{PSCM}: The canonical partially specified SCM bounds the query according to all possible FSCMs compatible with the dataset; 
    \textit{Independence}: Assumes all variables in $\vrz$ are independent;
    \textit{CBN}: For comparison, while not able to compute the query shown in the table, a CBN can compute the interventional query $p(\srpy=0 | \dop(\young=0), \gender=0, \makeup=1, \vvw = \xmapz_{\vvw}(\raisebox{-3pt}{\hbox{\includegraphics[height=11pt]{figs/87_real.png}}})) = 0.137$, which has a slightly different interpretation. Intervening on \young with the CBN resamples \gray and \glasses disregarding their observed values. Note that while the CBN value is contained in the PSCM interval, it is in this case in the opposite end of the interval compared to the FSCM value, suggesting the interventional query is not always a reliable approximation.\vspace{2mm}
    }
    \begin{tabular}{lc}
    \hline
    Model type &  $p(\srpy_{\dop(\young=0)}=0 | 
    \vvx=\raisebox{-3pt}{\hbox{\includegraphics[height=11pt]{figs/87_real.png}}}, 
    \spredy=1)$ \\
    \hline
    FSCM                &  0.121 \\
    PSCM (canonical)    & [0.121, 0.138] \\
    Independence        & 0.0 \\
    \hline
    \end{tabular}
    \label{tab:local_ps_young_per_model}
\end{table}

\begin{figure}[t]
    \centering
    \begin{tabular}{c@{}cc@{}c}
        \includegraphics[width=0.2\linewidth]{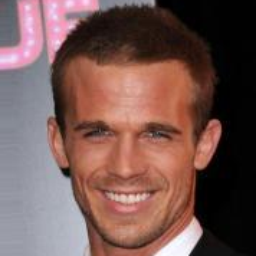} &
        \includegraphics[width=0.2\linewidth]{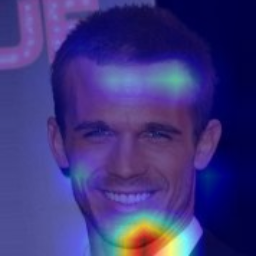} &
        \includegraphics[width=0.2\linewidth]{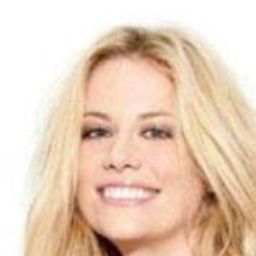} &
        \includegraphics[width=0.2\linewidth]{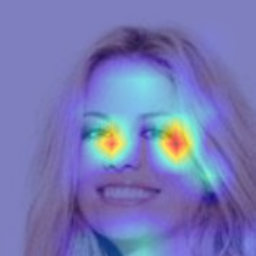}
    \end{tabular}
    \caption{
    The results of feature attribution by Grad-CAM on the original images in Figure \ref{fig:example_face_young}. 
    }
    \label{fig:grad_cam}
\end{figure}

\begin{figure}[t]
    \centering
    \begin{tabular}{c@{}cc@{}c}
        \includegraphics[width=0.2\linewidth]{figs/87_real.png} &
        \includegraphics[width=0.2\linewidth]{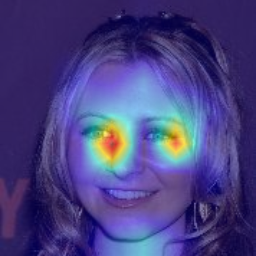} &
        \includegraphics[width=0.2\linewidth]{figs/87_old_fooled.png} &
        \includegraphics[width=0.2\linewidth]{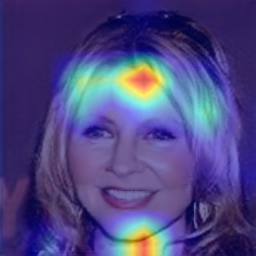}
    \end{tabular}
    \caption{
    The figure shows (from left to right) the original image from Figure \ref{fig:example_face_young_fooled}, Grad-CAM explanation of why the subject was classified as young, counterfactual image after $\dop(\young=0)$, Grad-CAM explanation of the erroneous classification that the subject is young.
    }
    \label{fig:grad_cam__error_case}
\end{figure}

While XAI systems that solely rely on visual assessment can be misleading \citep{Adebayo18}, saliency maps are still an often-used technique for attribution-based explanations of image classification systems. 
We therefore compare the results of our causal concept-based explanations with the results of Grad-CAM \citep{SelvarajuGradCAM17} using the implementation by \cite{jacobgilpytorchcam}. 
Consider first the results in Figure \ref{fig:grad_cam} which are intended to explain why the two actual (far left) images in Figure \ref{fig:example_face_young} were classified as being young.
Surprisingly, the chin area is important for the male's classification while the area around the eyes are of most importance for the classification of the female. 
Next, Figure \ref{fig:grad_cam__error_case} investigates the classification of the subject in Figure \ref{fig:example_face_young_fooled}. 
The first image was correctly classified as young, and according to the Grad-CAM explanation, the classifier has picked up on the area around the eyes to arrive at that conclusion. 
The explanation of the erroneously classified image to the right focuses on the forehead and the chin. 
Grad-CAM provides non-causal explanations, and therefore fails at giving counterfactual information. 
Instead of considering change, the explanation rather focuses on what part of the input space that are most indicative about the class being chosen the classifier.
Furthermore, since the explanations are pixel-based, we do not get insights regarding \textit{why} the highlighted areas are considered important. 

\paragraph{Global explanation}\ As an example of subgroup-explanations, we provide the probability that changing a single concept would suffice to change $h(\cdot)$'s prediction of a person from being young ($\srpy=1$) to being old in Table \ref{tab:singleton_ps_young}. 
We can write this probability compactly as $p(\srpy_{\dop(\intz=\intzv)}=0 | \intz=\vvz_s, \srpy=1)$, cf.\  Equation \ref{equ:subgroup_counterfactual}, where the conditioning on $\intz=\vvz_s$ is used to ensure that the value of $\vrz$ is indeed changed by the intervention. 
For simplicity we restrict ourselves to the most relevant singleton interventions, $|\intz| =1$.
As an example, consider the effect of setting $\intz=\{\gray\}$, $\vvz_s=0$ and $\intzv=1$. 
Now we are doing an investigation of the subpopulation of subjects that are classified as being young but do not have gray hair, looking for the counterfactual probability that the classifier would consider the subject to be old if their hair colour had been changed to gray. 
The relatively large value for this probability shows that hair colour is an important concept to explain the behaviour of the classifier, which may be used to spark further investigation into potential bias in the classifier before it can be deployed. 
We investigate this further in Table \ref{tab:gray_hair_subpop}, where the same calculation is broken down by gender and whether or not the subjects wear glasses. 
The effect is stronger among males than females, and albeit with a limited dataset, it also seems like the effect is stronger among those wearing glasses.

\delete{
\begin{table}[ht]
    \centering
    \begin{tabular}{lcc}
    \hline
    Intervention 
    &  $p(\srpy_{\dop(\intz=\intzv)}=0 | \vvz_s, \spredy=1)$ & TCAV($\srpy=0, \intz=\intzv$) \\
    \hline
    $\gray = 1$  & 0.318 & 0.960 \\
    $\glasses = 1$    & 0.002 & 0.990 \\
    $\makeup = 0$    & 0.000 & 0.930 \\
    $\young = 0$     & 0.972 & 1.000\\
    \hline
    \end{tabular}
    \caption{Probability of sufficiency for singleton interventions (i.e., $|\intz| =1$). Interpreted as an attribution score, it reflects the probability that the counterfactual intervention is sufficient to change the prediction from young to old. For comparison, concept attributions using TCAV scores \citep{kim2018interpretability} are included, which reflect the proportion of images predicted as young for which increasing concept activation will move the prediction towards old. Concept activations are identified with labelled data per concept, with no causal considerations. Here, TCAV identifies all concept interventions listed as moving the prediction towards old for more than $90\%$ of the images, suggesting comparable importance. Sufficiency scores however, directly reference the probability of crossing a decision boundary, with sound causal interpretation.}
    \label{tab:singleton_ps_young}
\end{table}
}

\begin{table}[ht]
    \centering
    \caption{
    Probability of sufficiency for singleton interventions (i.e., $|\intz| =1$). Interpreted as an attribution score, it reflects the probability that the counterfactual intervention is sufficient to change the prediction from young to old. 
    }
    \begin{tabular}{lc}
    \hline
    Intervention 
    &  $p(\srpy_{\dop(\intz=\intzv)}=0 | \vvz_s, \spredy=1)$ \\
    \hline
    $\gray = 1$     & 0.318 \\
    $\glasses = 1$  & 0.002 \\
    $\makeup = 0$   & 0.000 \\
    $\young = 0$    & 0.972\\
    \hline
    \end{tabular}
    \label{tab:singleton_ps_young}
\end{table}

\begin{table}[ht]
    \centering
    \caption{Probability of sufficiency for \gray, broken down on each subject's \gender and status regarding \glasses. The \textit{Count}-column shows the number of observations in the test-set (out of $2000$) that met the requirements. There were, for instance, 43 males without gray hair wearing glasses that were initially classified as young in the test-set.\vspace{2mm}}
    \begin{tabular}{c|c|r|c}
    \hline 
    \multicolumn{3}{c|}{Subgroup} & \multirow{2}{*}{$p(\srpy_{\dop(\intz=\intzv)}=0 | \vvz_s, \spredy=1)$} \\ \cline{1-3}
    $\gender$ & $\glasses$    & Count           &       \\ \hline 
    Male   & Both       &  519\phantom{.} & 0.580 \\
    Male   & True       &   43\phantom{.} & 0.814 \\
    Male   & False      &  476\phantom{.} & 0.559 \\ \hline
    Female & Both       &  986\phantom{.} & 0.181 \\
    Female & True       &   15\phantom{.} & 0.400 \\
    Female & False      &  971\phantom{.} & 0.177 \\ \hline
    Both   & Both       & 1505\phantom{.} & 0.318 \\ \hline
    \end{tabular}
    \label{tab:gray_hair_subpop}
\end{table}

Since setting $\intz=\{\young\}$ implies $\infz\neq\emptyset$, we must again marginalize out the uncertainty over $\infz$. 
How this is done depends on how the causal model $M$ is represented, cf.\ Table \ref{tab:ps_young_per_model}. For this particular query, the canonical interval is narrow, such that the choice of FSCM has limited influence on the calculated probability of sufficiency. Using a model with independent concept variables slightly underestimates the global effect of the intervention, as was also shown to be the case locally for the image in Figure \ref{fig:example_face_young_fooled}.

\begin{table}[ht]
    \centering
    \caption{
    Probability of sufficiency for counterfactual intervention on the variable $\intz = \young$. 
    \textit{FSCM}: The FSCM is the only model representation able to calculate the counterfactual probabilities directly; 
    \textit{PSCM}: The canonical PSCM bounds the query according to all possible FSCMs compatible with the dataset;  
    \textit{Independence}: Assumes all variables in $\vrz$ are independent. \vspace{2mm}
    }
    \begin{tabular}{lc}
    \hline
    Model type &  $p(\srpy_{\dop(\young=0)}=0 | \young=1, \srpy=1)$ \\
    \hline
    FSCM   & 0.972 \\
    PSCM (canonical) & [0.970, 0.973] \\
    Independence & 0.966 \\
    \hline
    \end{tabular}
    \label{tab:ps_young_per_model}
\end{table}

Global explanations are shown for the attractiveness classifier in Tables \ref{tab:singleton_ps_attractive} and \ref{tab:ps_makeup_per_model}. 
Table \ref{tab:singleton_ps_attractive} shows the effect of singleton interventions in young people, grouped by their gender.  
The numbers are understood as follows, using the value 0.139 in the first row as an example: This is the probability that a female ($\gender=0$) currently 
classified as unattractive ($\spredy=0$) not having \concept{5 o'Clock Shadow} ($\intz=\neg\intzv=0)$ would be classified as attractive ($\spredy=1$) if she did ($\intz=1$). 
The table indicates that females improve their chance of being classified as attractive the most if they remove a possible \concept{Big Nose}, whereas the best strategy for men is to obtain a \concept{Pointy Nose} if they do not already have one. 
For comparison, concept attributions using TCAV scores \citep{kim2018interpretability} are included in Table \ref{tab:singleton_tcav_attractive}.
These values reflect how concept interventions are moving the prediction towards the subject being seen as more attractive.
According to the TCAV scores, obtaining \concept{Arched Eyebrows} is most beneficial for both genders, followed by the removal of \concept{Bushy Eyebrows}.  
The differences in attribution from the two approaches are to be expected. 
TCAV concept directions are identified with labelled data per concept, with no causal considerations. 
Sufficiency scores however, directly reference the probability of crossing a decision boundary, with sound causal interpretation. As an example, the sufficiency attributions for \concept{Arched Eyebrows} in Table \ref{tab:singleton_ps_attractive} are estimated based on controlled generation of data where only the eyebrows change, thus reflecting the isolated effects of this change. The TCAV scores using observed data (conditioning on a value for \concept{Arched Eyebrows}) are more prone to influence from correlated concepts. This could be an explanation for the contrasting results obtained for this concept by the two approaches. 

Note that the sufficiency scores also better reflect subgroup tendencies, with the effect of interventions to e.g. \concept{Big Nose}, \concept{Arched Eyebrows} and \concept{Smiling} being more distinct when comparing the male and female subgroups, than what is reflected by the TCAV scores.

\begin{table}[ht]
    \centering
    \caption{Probability of sufficiency for singleton interventions to leaf variables in the explanation graph in Figure \ref{fig:example_attractive}, conditioned on \young = 1 and \gender, for changing the output of the attractiveness classifier from not attractive to attractive. \vspace{2mm}}
    \begin{tabular}{llcc}
    \hline
    \multicolumn{2}{l}{Intervention 
    }
    &  \multicolumn{2}{l}{$p(\srpy_{\dop(\intz=\intzv)}=1 | \intz=\neg\intzv, \young=1, \gender, \spredy=0)$} \\
    \hline
    $\intz$& $\intzv$ & \gender = 0 & \gender = 1 \\
    \hline
    \multirow{2}{*}{\concept{5 o'Clock Shadow}} & 1  & 0.139 &  0.117 \\
     & 0  & - &  0.000 \\
    \hline 
    \multirow{2}{*}{\concept{Arched Eyebrows}} & 1 & 0.034 & 0.003\\
     & 0  & 0.085 & 0.214\\
    \hline
    \multirow{2}{*}{\concept{Bushy Eyebrows}} & 1    & 0.177  & 0.109 \\
    & 0     & 0.000 & 0.000\\
    \hline
    \multirow{2}{*}{\concept{Big Nose}} & 1    & 0.000  & 0.004 \\
     & 0     & 0.574 & 0.147\\
    \hline
    \multirow{2}{*}{\concept{Pointy Nose}} & 1    & 0.350 & 0.306 \\
    & 0    & 0.000 & 0.000 \\
    \hline
    \multirow{2}{*}{\smiling} & 1   & 0.242 & 0.181 \\
     & 0   & 0.000 & 0.009 \\
    \hline
    \end{tabular}
    \label{tab:singleton_ps_attractive}
\end{table}

\begin{table}[ht]
    \centering
    \caption{TCAV scores quantifying the importance of various concepts for classifying attractiveness. Concept directions are probed at the penultimate convolutional layer, as is commonly done due to later layers having the highest linear separability. The held-out background dataset $\dset^*$ is limited to samples classified as not attractive and is further partitioned based on \gender and intervened concepts. For $\intzv = 0$, we flip the concept direction to $-\vv{v}_\vrz$, simulating removal of the concept.\vspace{2mm}}
    \begin{tabular}{llcc}
    \hline
    \multicolumn{2}{l}{Intervention 
    }
    & \multicolumn{2}{l}{
    $\text{TCAV}_{\ell}(\spredy=1, 
    \intzv, 
    \dset_{\spredy=0 \cap \intz=\neg \intzv \cap \gender})$} \\
    \hline
    $\intz$ & $\intzv$ & \gender = 0 & \gender = 1 \\
    \hline
    \multirow{2}{*}{\concept{5 o'Clock Shadow}} & 1  & 0.256 &  0.251 \\
     & 0  & - &  0.788 \\
    \hline 
    \multirow{2}{*}{\concept{Arched Eyebrows}} & 1 & 0.995 & 0.977\\
     & 0  & 0.011 & 0.011\\
    \hline
    \multirow{2}{*}{\concept{Bushy Eyebrows}} & 1    & 0.182  & 0.292 \\
    & 0     & 0.835 & 0.796\\
    \hline
    \multirow{2}{*}{\concept{Big Nose}} & 1    & 0.314  & 0.270 \\
     & 0     & 0.745 & 0.714 \\
    \hline
    \multirow{2}{*}{\concept{Pointy Nose}} & 1    & 0.660 & 0.646 \\
    & 0    & 0.325 & 0.344 \\
    \hline
    \multirow{2}{*}{\smiling} & 1   & 0.642 & 0.588 \\
     & 0   & 0.408 & 0.443 \\
    \hline
    \end{tabular}
    \label{tab:singleton_tcav_attractive}
\end{table}

\begin{table}[ht]
    \centering
    \caption{Probability of sufficiency for counterfactual intervention $\dop(\makeup=1)$ for the subgroup \young=1, \gender=0 and \makeup=0 classified as not attractive. Values for FSCM, PSCM and Independence calculated as detailed for Table \ref{tab:ps_young_per_model}.\vspace{2mm}}
    \scalebox{.9}{
    \begin{tabular}{lc}
    \hline
    Model type &  $p(\srpy_{\dop(\makeup=1)}=1 | \young=1,\gender=0,\makeup=0, \srpy=0)$ \\
    \hline
    FSCM   & 0.410 \\
    PSCM (canonical) & [0.401,0.442] \\
    Independence & 0.353 \\
    \hline
    \end{tabular}
    }
    \label{tab:ps_makeup_per_model}
\end{table}

Lastly, Table \ref{tab:ps_makeup_per_model} is included as an example of an explanation based on an intervention to a non-leaf variable in the explanation graph for the attractiveness classifier. For young females without makeup, the intervention \makeup=1 is considered. It is shown that accounting for the causal influence of \makeup on the concepts representing appearance of \concept{Pointy Nose} and \concept{Arched Eyebrows}, attributes a higher effect of \makeup on the attractiveness prediction, as measured by the probability of sufficiency, compared to the independent intervention. The probability of sufficiency for \concept{Pointy Nose}=1 in particular, shown in Table \ref{tab:singleton_ps_attractive}, suggests that \concept{Pointy Nose}=1 increases the chance of predicted attractiveness, which is accounted for by the FSCM-based calculation of the probability of sufficiency of \makeup=1. 
Independent intervention on \makeup however, will not influence the value of \concept{Pointy Nose}, and so the appearance of a pointy nose in the generated image will not change. The PSCM interval included in Table \ref{tab:ps_makeup_per_model} further suggest the independence-model underestimates the true causal effect of \makeup compared to all possible FSCMs compatible with the data, with the independent attribution notably lower that that of the lower PSCM bound.








\subsection{Discussion}\label{sec:experiment_discuss}



\subsubsection{Assumptions}\label{sec:example_discussion_assumptions}

Section \ref{sec:results} presents a set of results obtained by implementing and testing a proof-of-concept explanation model for classifiers trained on the CelebA dataset. As discussed in Section \ref{sec:intro}, understandability and 
fidelity of explanations are contingent on the viability of the model assumptions under which explanations are generated. Thus, this section discusses the assumptions made for the proof-of-concept model presented, and their implications regarding explanation interpretation. While the design of this model is intentionally simplistic in order to feasibly illustrate the potential of the general framework, implementation-specific weaknesses are discussed to highlight challenges that may arise when model assumptions are violated.

\paragraph{The explanation vocabulary} The CelebA dataset includes labels for 40 binary attributes. In order to feasibly model the SCM over the explanation vocabulary $\vrz$, a subset of these 40 attributes is chosen for $\vrz$, as detailed in Section \ref{sec:ex_models}. The remaining attributes are considered part of $\vrw$. Two assumptions are made for this to be meaningful. The first is that the attributes included in $\vrz$ are independent of those left to be encoded in $\vrw$, such that the effects of interventions in $\vrz$ are reliably mapped to $\vrx$ via $\zmapx$. The second is that the selection of concepts for $\vrz$ is reasonable for explaining $h$ in areas of interest. As this is typically the decision boundary of $h$, this means that $\vrz$ should include those concepts that, when intervened upon, will move the corresponding input across this boundary (locally or globally). This selection should however not be influenced by expectations regarding which concepts should ideally guide $h$, as such a restriction for $\vrz$ could prevent the framework from detecting unexpected and potentially unwanted behaviour in $h$.

\paragraph{The causal model} Given a selection of variables $\vrz$ now assumed to be independent of all other information present in $\vrx$, the assumption that the causal structure and the structural equations of the SCM $M$ accurately reflect the world, concludes the vocabulary context. The causal structure in the CelebA examples are mainly learned from data, by detecting (conditional) dependence among pairs of variables and querying an expert (an LLM or the model designer) for the most likely causal direction of this dependence. This results in substructures such as \gender $\rightarrow$ \concept{Smiling}, which could be argued not a true causal process. 
While a person's gender is not expected to have direct causal influence on the likelihood of them smiling, the dependence in the data is more likely a result of either an indirect effect through a mediator, such as a person's smiling habits being a result of the person's previous experiences with their society's gender biased expectations, or a confounder, such as the data selection process. 
If the former is the case, it is reasonable to keep the simplified causal component as part of the causal structure, omitting lower level details regarding the unobserved mediating factor. 
This argument is made in favour of including the components \gender $\rightarrow$ \concept{Smiling} and \gender $\rightarrow$ \makeup in the example. 
If the latter is the case, one might wish to enforce independence in the final causal structure used for explanation, or otherwise adapt the causal model to include the unobserved confounder. 
In the discussed examples, the variables \young and \gender are modelled as independent despite dependence present in the data, according to such an argument. As the final choice of causal structure determines how explanations should be interpreted, this highlights the importance of including expert knowledge when designing the causal model.

Finally, when designing the structural equations of the model, reliable expert knowledge is again important. The PSCM intervals included illustrate that data and causal structure are not sufficient for exact counterfactual calculation, and knowledge about causal relationships may improve upon this. 

\paragraph{StarGAN concept interventions}
The StarGAN generator implemented to represent $\zmapx$ takes an image $\vvx$ (consistent with concepts $\vvz$) and a concept vector $\vvz'$ as input, and generates $\vvx'$ by minimal alterations to original $\vvx$ such that $\vvx'$ is consistent with $\vvz'$. This domain translation is learned by optimising over the ability of a set of concept classifiers to correctly classify generated samples, including a reconstruction loss encouraging $\zmapx(\vvx', \vvz) = \vvx$. Thus, $\vrw = \vrx$ which violates the framework assumption $\vrw \ind \vrz$. It is rather assumed that the design of $\zmapx$ ensures that when generating $\vvx' = \zmapx(\vvx, \vvz')$, all information in $\vvx$ pertaining to original concept values $\vvz$ is replaced by introduced $\vvz'$, without influence from $\vvz$. This assumption is validated to the extent of inspection of generated samples. 

The focus of the implemented generator is efficient and reliable modelling of concept interventions, leaving state-of-the-art generative quality for future work. While the general quality of the generated images, as illustrated in Figures \ref{fig:example_face_young} and \ref{fig:example_face_young_fooled}, may have unintended influence on the presented results, this is assumed negligible for the purpose of this proof of concept example. 

The generator is learned from observations, and is therefore not guaranteed to isolate concepts as intended. For instance, if $\vrz = \{\young\}$, it is plausible that the generator may learn to include gray hair as a data instance is mapped from young to old, as these attributes are expected to be dependent. However, if $\vrz = \{\young, \gray\}$, the generator is incentivised to avoid confusing the two attributes, as classifiers predicting both attributes influence learning. 
Because it is assumed that the explanation vocabulary is independent of all other information in $\vrx$, it is therefore assumed that a single StarGAN generator trained with the complete vocabulary vector $\vrz$ as input, will both distinguish these concepts successfully, as well as avoid that any concept retains information expected to be part of $\vrw$. This is verified by inspection.

Finally, some of the attributes of the CelebA dataset are highly simplistic, binarized versions of complex concepts, an example of this being the attribute \young representing a person's age as either young or old. Data generation is influenced accordingly. While the original dataset contains people of varied age, generated data are mapped to either young or old as learned binary categories. If an instance $\vvx$ representing a young person is input to the generator with a concept vector where \young=1 is retained, the expectation is that the age of the generated image will remain unaltered. However, 
the generator will map the input to the binary representation of young, such that a very young person may be aged, while a close to old person will appear younger in the generated counterpart. Note that in the latter case, this can be especially adverse when explaining misclassified instances.
Generated explanations for a person who is close to old may wrongly attribute a change in prediction to a change to a given concept, while it is the slight, unintended age adjustment that ensures the decision boundary is crossed. The results presented in Section \ref{sec:results} again assume this effect to be negligible for the data instances chosen to be explained. As this assumption is expected to be violated in a more general context, it illustrates the challenge of ensuring that explanations are generated and interpreted in the appropriate context. 



\paragraph{Dataset}

In the examples presented here, both the model $h$ to be explained and the models that generate the explanations $\zmapx$ and (parts of) $M$ are learned from the same dataset. While this is not uncommon for XAI, it is briefly discussed as a potential cause of explanation bias.

The ability to explain decisions is important for many reasons, one of which is to detect unintended behaviour of $h$, as can be a result of learning $h$ from spurious correlations present in biased datasets. In this case, learning $\zmapx$ and $M$ from the same dataset could hinder detection of this particular behaviour. Considering two concepts that are expected to be independent, but are in fact correlated in the data by some unknown confounder in the data generation process. If during vocabulary selection, one of these concepts are included in $\vrz$ while the other is left in $\vrw$, the generator may learn a joint version of these concepts while referencing this information by name of the concept in $\vrz$ only. This highlights the importance of a sufficiently extensive vocabulary. 

When learning the causal model $M$, it is typically a combination of data and expert knowledge that decides the final model, such that spurious correlations are more easily detected and removed if present among concepts at this level. For instance, there is a correlation between attributes age and gender in the CelebA dataset. This is identified in the presented example as spurious, such that these two concepts are modelled as independent in the explanation model considered. The ability to incorporate this type of expert knowledge is an important advantage of the causal framework. However, whenever data guides the learning, the causal model risks being biased by spurious correlations similarly to $h$ and $\zmapx$.

\subsubsection{Choice of abstraction of model}
\label{sec:model_abstraction}

The level of abstraction of a causal model $M$ determines how the total causal influence of one variable on another is divided between direct and indirect causal influence. If the explanation vocabulary is restricted to be $\vrz = \{\young\}$, the causal structure considered is 
 \begin{equation*}
     \young \rightarrow \vrx \rightarrow \srpy
 \end{equation*}
modelling the total causal effect as direct effect, abstracting away any mediators.

The example presented in Figure \ref{fig:example_young}, considers the structure

\begin{center}
\begin{tikzpicture}
\node[] (y) at (-2.5,0.2) {};
\node[] (x) at (2.1,0.2) {};
\node[draw=none] (e) at (0,0)
 {$\young \rightarrow \{\gray, \glasses\} \rightarrow \vrx \rightarrow \srpy$};
 \draw[a2] (y) to [out=17, in=163] (x);
\end{tikzpicture}
\end{center}

\noindent
such that some of the total effect is mediated through variables gray hair and glasses. 

With the data being images of people limited to visual information, it is conceivable that the vocabulary could be extended further such that all causal effect from young to an image $\vvx$ is mediated through some physical feature explicit in $\vrz$:

 \begin{equation*}
     \young \rightarrow \text{all physical features} \rightarrow \vrx \rightarrow \srpy
 \end{equation*} 

In all three models, the total causal influence of young on the image and prediction is the same, illustrating that it is a reliable measure independent of abstraction. However, if an explanation model ignores the causal structure by intervening on variables as if independent, only the direct effect is captured, which varies from equal to the total effect for the first model, to 0 for the last model. Thus, direct influence is not a consistent measure.



\section{Discussion}

The framework for causal post-hoc XAI presented in this paper constitutes a substantial extension to a model $h$ to be explained, requiring several additional model components and more extensive datasets than those used to train $h$. While implementation is more demanding than feature-based or non-causal XAI methods, the main argument in favour of this framework is the understandability it facilitates, both in terms of vocabulary and causal interpretation. 
The concept vocabulary requires concept information such as labels, but allows explanation of the model in an extended context in which changes to the model input, and subsequently output, has meaningful interpretation. The causal modelling additionally requires expert knowledge, but lets the framework admit counterfactual reasoning within the extended context in order to identify causes of model outputs, a property identified as especially advantageous for human explainees \citep{miller19}. As discussed in Section \ref{sec:model_abstraction}, having a causal model over the considered vocabulary allows estimating the total causal effect between variables, argued a more consistent measure than direct causal effect. 
This illustrates how even though causal modelling is required for generation of explanations, interpretation of the resulting explanations will require less awareness of model details.  

Section \ref{sec:example_discussion_assumptions} discusses the challenges arising when model assumptions are violated, which suggests that model choices should be made carefully. Whenever model assumptions are found to introduce a bias in explanations, either the model must be adapted or, if not feasible, the explainee must be informed about the biased context in which the explanations are valid. This is however not meant to indicate that complex models with less assumptions are always favoured over simpler ones. For instance, to ensure that explanations are generated in accordance with their expected causal interpretation, the causal model is primarily required to reflect a level of abstraction consistent with that of the explainee's causal understanding of the world (assumed imprecise but not false). With understandability as a main goal, abstracting away certain low level singular causal mechanisms is preferable, condensing relevant information into a more accessible representation. 

Another significant advantage of the causal explanation model discussed in Section \ref{sec:example_discussion_assumptions} is that the process of causal modelling of concepts, reliant on some contribution of expert knowledge, is more robust regarding detecting and removing spurious correlations than data-driven modelling. Thus, the explanation model can identify spurious relationships learned by the predictor, even in some cases where the same dataset is used to learn the explanation model. With expert knowledge becoming more accessible through LLMs, as demonstrated with the example presented, this has potential particularly for XAI targeting model improvement through detecting unintended model behaviour. 

The part of the explanation model that is the most challenging to design is typically the structural equations of the FSCM. Canonical PSCM intervals for counterfactual queries are included in the results presented in Section \ref{sec:results} to illustrate what is lost when the true causal mechanisms cannot be determined. For some queries, as seen in the results, the intervals are narrow, suggesting a reasonable approximation of the true value may be given by the interval mean or even by approximate queries using a CBN, as exact model choice is of less consequence. 
However, this is not guaranteed to be the case. In general, when conditioning on observed events of low probability, a counterfactual probability may range from 0 to 1 depending on the model considered. Two different FSCMs that are both consistent with the data may therefore generate opposite explanations of the same model, and even claim these as certain outcomes. Considering that less likely observations can be especially important to explain reliably, for instance when making sure predictions are meaningful even in areas with limited training data, the general framework still requires access to an FSCM to guarantee arbitrary query calculation. Note that non-canonical PSCMs defined by partial expert knowledge may reduce the size of the intervals, extending the scope of queries that may be approximated. Future work may further compare the feasibility and capacity of various approaches to exact and approximate calculations. 

While the types of explanations discussed in this work are based on calculating the probability of sufficiency of concept interventions, this is not a limitation of the framework, and future work on extending the types of explanations considered may explore probabilities of necessity, as well as considering explanations exploring actual causation in line with \citep{beckers22}. 

Other lines of future work include developing general architectures for modelling the concept-to-data generative process, respecting the requirements detailed for the concept space while improving generative quality. Automated selection and detection of relevant concepts for the explanation vocabulary may be further explored by inspection of the model to be explained prior to designing the causal model for explanation. In a similar vein, comparing concept attributions can guide the selection of a reasonable vocabulary from a larger set of predefined concepts. Techniques for concept detection such as sparse autoencoders, that have been shown to be useful for detecting interpretable concepts without access to human-annotated data \citep{cunningham2023sparse}, may further help suggest relevant concepts to be included in the vocabulary without being influenced by preconceptions about concept importance. Automated concept detection could help ensure that the vocabulary contains the most relevant concepts by considering how well they approximate the original model, and whether interventions to the concepts are likely to cross a decision boundary.


Finally, existing concept attribution methods may be used to guide the search for selecting optimal explanations. Specifically, the number of counterfactuals that can be considered grows exponentially in the number of concepts, for instance $2^{|\vrz|}$ for binary concepts. In those computationally heavy cases, concept attribution can serve as a heuristic, for example, by ordering concepts to intervene on, combined with early stopping strategies. 





\section{Conclusion}

This paper introduces a conceptual framework for causal concept-based XAI, that is designed to allow generation of causal and counterfactual explanations in a concept vocabulary. 
The framework consists of a causal model over the concepts in the vocabulary, and a causal generator that maps concepts to data features. 
The generation of local and global explanations using this framework, based on the probability of sufficiency of concept interventions, is detailed. 
A proof of concept explanation model is applied to classifiers trained on the CelebA dataset, and a selection of explanations generated is presented. 
These results show promise, and motivate future work towards improved realisation of the conceptual framework in generalised domains.

\appendix
\renewcommand{\thesection}{Appendix \Alph{section}}

\section{}
\label{sec:FSCMdetails}

This section describes the details of the modelling of structural functions for the SCMs used for calculating the results presented in Section \ref{sec:results}, and assumes discrete variables.  

A PSCM $\langle\vru, \vrz, \mathcal{F}\rangle$ consistent with a given causal graph $\mathcal{G}$ is considered first. For each concept $\srz_i \in \vrz$ there is a function $f_i \in \mathcal{F}$ such that $f_i(\mathbf{pa}_i,\sru_i) = \srz_i$, where $\mathbf{pa}_i$ denotes the set of endogenous parents of variable $\srz_i$ in $\mathcal{G}$ and $\sru_i$ is the exogenous parent of $\srz_i$. Each function is defined as $f_i(\mathbf{pa}_i,\sru_i)  = f_{i, u_i}(\mathbf{pa}_i)= \srz_i$, such that a value $u \in \Omega_{\sru_i}$ indexes a distinct deterministic function $f_{i,u}: \Omega_{\mathbf{pa}_i} \rightarrow \Omega_{\srz_i}$. There are a total of $|\Omega_{\srz_i}|^{|\Omega_{\mathbf{pa}_i}|}$ possible functions. Thus, for a canonical PSCM it is the case that $|\Omega_{\sru_i}|=|\Omega_{\srz_i}|^{|\Omega_{\mathbf{pa}_i}|}$. A non-canonical PSCM is defined by selecting a subset of the canonical set of functions to be allowed indexed by the exogenous variable, based on domain knowledge. 

Given a dataset $\dset$, a PSCM will correspond to a credal network \citep{zaffalon2020structural, zaffalon2024efficient}, such that for each exogenous variable $\sru_i$, there is a set of probability distributions that are consistent with $\dset$, denoted $\mathcal{K}(\sru_i)$. Thus, the PSCM can be seen as a set of FSCMs as defined by the Cartesian product $\times_i \mathcal{K}(\sru_i)$. Intervals for counterfactuals are calculated bounding the query according to this set of FSCMs, and the method introduced in \citep{bjoru2024, bjoru2025ijar, bjoru2025semimarkovian} is applied for the interval calculations presented in Section \ref{sec:results}.

\begin{table}[t]
    \centering
    \caption{The structural function and exogenous distribution chosen for variable \gray. \vspace{2mm}}
    \begin{tabular}{ccc}
    \hline
        $u \in \Omega_{\sru_{\gray}}$ & $f_{\gray, u}(\gender, \young) = \gray$ & $p(u)$\\
        \hline
         $u_0$ & 0 & 0.774\\
         $u_1$ & $\gender \land \young$ & 0 \\
         $u_2$ & $\gender \land \neg\young$ & 0.151 \\
         $u_3$ & $\gender$ & 0 \\
         $u_4$ & $\neg\gender \land \young$ & 0 \\
         $u_5$ & $\young$ & 0 \\
         $u_6$ & $\gender \oplus \young$ & 0 \\
         $u_7$ & $\gender \lor \young$ & 0 \\
         $u_8$ & $\neg\gender \land \neg\young$ & 0 \\
         $u_9$ & $\gender \oplus \neg\young$ & 0 \\
         $u_{10}$ & $\neg \young$ & 0.071\\
         $u_{11}$ & $\gender \lor \neg\young$ & 0.003 \\
         $u_{12}$ & $\neg\gender$ & 0\\
         $u_{13}$ & $\neg\gender \lor \young$ & 0 \\
         $u_{14}$ & $\neg\gender \lor \neg\young$& 0  \\
         $u_{15}$ & 1 & 0.001 \\
    \hline
    \end{tabular}
    \label{tab:structuralfunctionex}
\end{table}

The FSCMs used for the examples are defined by designing a non-canonical PSCM such that the set $\times_i \mathcal{K}(\sru_i)$ contains exactly one distribution $p(\vru)$ consistent with the data $\dset$. This is exemplified by detailing the model component $\gender \rightarrow \gray \leftarrow \young$ from the graph shown in Figure \ref{fig:example_young}, and the structural function $f_{\gray}$. The size of the canonical domain of exogenous $\sru_{\gray}$ is given by $|\Omega_{\sru_{\gray}}|=2^{{2\cdot2}}$, i.e. there are 16 possible functions to consider. These are listed in Table \ref{tab:structuralfunctionex}. The functions corresponding to exogenous values $u$ for which $p(u) = 0$ are not permitted by the FSCM. The probability of the remaining functions are given by the data.

The model component $f_{\gray}, p(\sru_{\gray})$ defined in Table \ref{tab:structuralfunctionex} is designed based on simple assumptions, such as eliminating functions for which a person will have gray hair if young and not if old, while ensuring the restricted model fits the data. This approach is repeated for all the structural functions of the FSCMs used for the example results, and the resulting FSCMs should be interpreted as reasonable model suggestions given the data provided. For instance, the low number of gray haired individuals in the CelebA dataset is reflected in the exogenous distribution presented in Table \ref{tab:structuralfunctionex}, which suggest that 75\% of the population considered will never get gray hair, regardless of age and gender. For explanations more in line with common expectations about the occurrence of gray hair in the general population, an alternative approach relying more heavily on domain knowledge could rebalance the distribution $p(\sru_{\gray})$ accordingly.

\bibliographystyle{plainnat}
\bibliography{references}

\end{document}